\documentclass{article}

\usepackage{microtype}
\usepackage{graphicx}
\usepackage{booktabs} 

\usepackage{hyperref}

\usepackage{times}
\usepackage{epsfig}
\usepackage{graphicx}
\usepackage{amsmath}
\usepackage{amssymb}
\usepackage{color}
\usepackage{colortbl}
\usepackage{balance}
\usepackage{soul}

\usepackage{gensymb}
\usepackage{booktabs}
\usepackage{graphicx}

\usepackage{subcaption}
\usepackage{dblfloatfix}
\usepackage{setspace}
\usepackage{bbm}
\usepackage{multirow}

\newcommand{\ctext}[3][RGB]{%
  \begingroup
  \definecolor{hlcolor}{#1}{#2}\sethlcolor{hlcolor}%
  \hl{#3}%
  \endgroup
}

\renewcommand{\paragraph}[1]{\noindent\textbf{#1}}

\newcommand{\name}{CenterNet2 }

\definecolor{lightgray}{RGB}{238,238,236}

\newcommand{\shortrefeq}[1]{Eq.~\eqref{eq:#1}}

\newcommand{\lblsec}[1]{\label{sec:#1}}
\newcommand{\lbleq}[1]{\label{eq:#1}}

\definecolor{lightgray}{rgb}{0.83, 0.83, 0.83}

\usepackage{hyperref}
\usepackage[accepted]{icml2021}

\icmltitlerunning{Probabilistic two-stage detection}

\begin{document}

\twocolumn[
\icmltitle{Probabilistic two-stage detection}

\begin{icmlauthorlist}
\icmlauthor{Xingyi Zhou}{ut}
\icmlauthor{Vladlen Koltun}{intel}
\icmlauthor{Philipp Kr{\"a}henb{\"u}hl}{ut}
\end{icmlauthorlist}

\icmlaffiliation{ut}{UT Austin}
\icmlaffiliation{intel}{Intel Labs}

\icmlcorrespondingauthor{Xingyi Zhou}{zhouxy@cs.utexas.edu}

\icmlkeywords{Computer vision, object detection}

\vskip 0.3in
]

\printAffiliationsAndNotice{} 

\begin{abstract}

We develop a probabilistic interpretation of two-stage object detection. We show that this probabilistic interpretation motivates a number of common empirical training practices. It also suggests changes to two-stage detection pipelines. Specifically, the first stage should infer proper object-vs-background likelihoods, which should then inform the overall score of the detector. A standard region proposal network (RPN) cannot infer this likelihood sufficiently well, but many one-stage detectors can. We show how to build a probabilistic two-stage detector from any state-of-the-art one-stage detector. The resulting detectors are faster and more accurate than both their one- and two-stage precursors. Our detector achieves 56.4 mAP on COCO test-dev with single-scale testing, outperforming all published results. Using a lightweight backbone, our detector achieves 49.2 mAP on COCO at 33 fps on a Titan Xp, outperforming the popular YOLOv4 model.

\end{abstract}

\section{Introduction}

Object detection aims to find all objects in an image and identify their locations and class likelihoods~\cite{girshick2014rich}.
One-stage detectors jointly infer the location and class likelihood in a probabilistically sound framework~\cite{lin2018focal,liu2016ssd,redmon2017yolo9000}.
They are trained to maximize the log-likelihood of annotated ground-truth objects, and predict proper likelihood scores at inference.
A two-stage detector first finds potential objects and their location~\cite{uijlings2013selective,zitnick2014edge,ren2015faster} and then (in the second stage) classifies these potential objects.
The first stage is designed to maximize recall~\cite{ren2015faster,he2017mask,cai2018cascade}, while the second stage maximizes a classification objective over regions filtered by the first stage.
While the second stage has a probabilistic interpretation, the combination of the two stages does not.

In this paper, we develop a probabilistic interpretation of two-stage detectors.
We present a simple modification of standard two-stage detector training by optimizing a lower bound to a joint probabilistic objective over both stages.
A probabilistic treatment suggests changes to the two-stage architecture.
Specifically, the first stage needs to infer a calibrated object likelihood.
The current region proposal network (RPN) in two-stage detectors is designed to maximize the proposal recall, and does not produce accurate likelihoods.
However, full-fledged one-stage detectors can.

\begin{figure}[t]
\begin{center}
\includegraphics[width=0.95\linewidth,page=4]{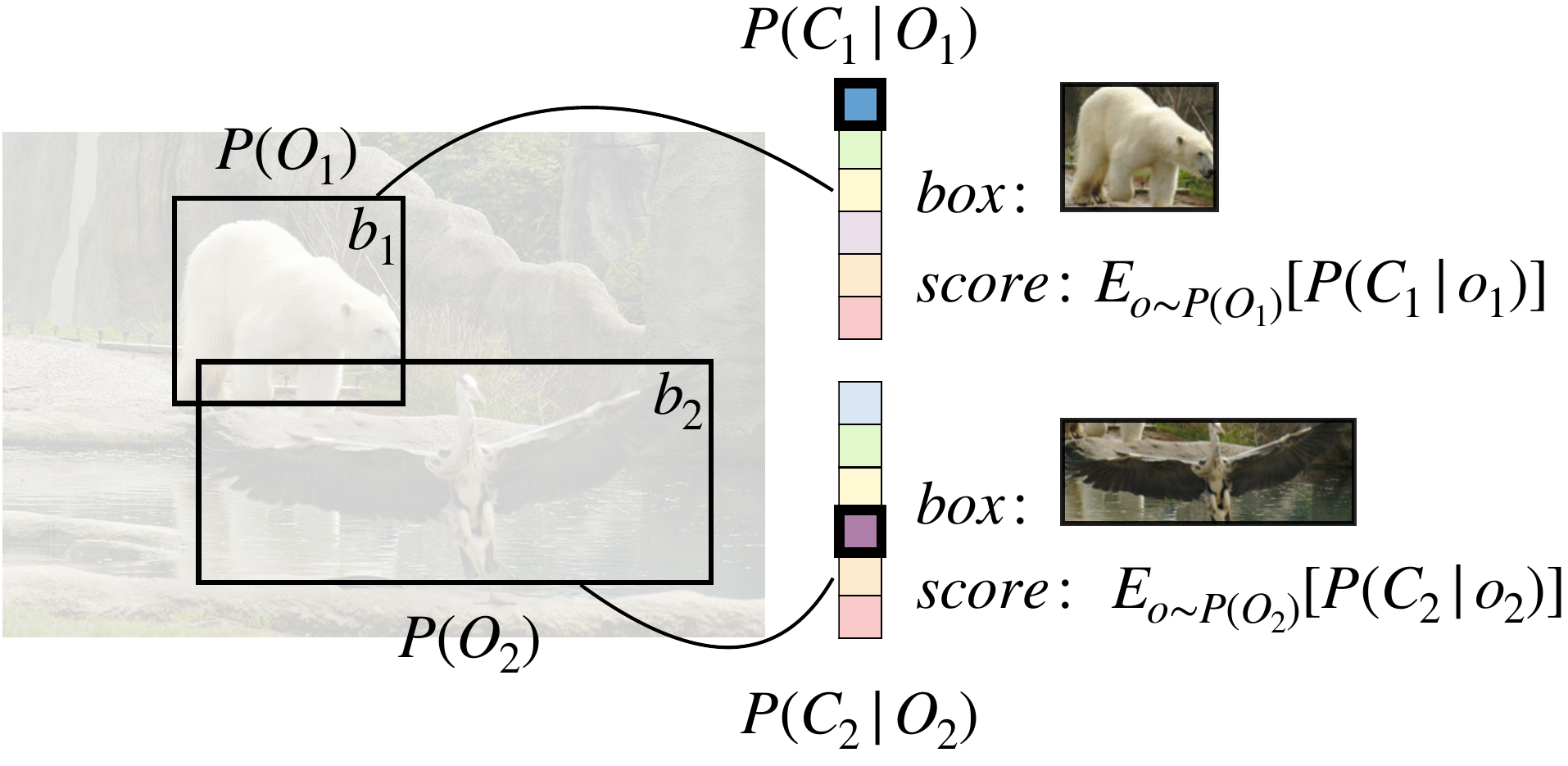}
   \begin{subfigure}[b]{0.5\linewidth}\caption{First stage:\\Object likelihood}\label{fig:teaser_fs}\end{subfigure}%
   \begin{subfigure}[b]{0.5\linewidth}\caption{Second stage: \\Conditional classification}\label{fig:teaser_ss}\end{subfigure}%
\end{center}
\vspace{-5mm}
   \caption{
   Illustration of our framework. A class-agnostic one-stage detector predicts object likelihood. A second stage then predicts a classification score conditioned on a detection.
   The final detection score combines the object likelihood and the conditional classification score.
   }
\vspace{-6mm}
\label{fig:framework}
\end{figure}

We build a probabilistic two-stage detector on top of state-of-the-art one-stage detectors.
For each one-stage detection, our model extracts region-level features and classifies them.
We use either a 
Faster R-CNN~\cite{ren2015faster} or a cascade classifier~\cite{cai2018cascade} in the second stage.
The two stages are trained together to maximize the log-likelihood of ground-truth objects.
At inference, our detectors use this final log-likelihood as the detection score.

A probabilistic two-stage detector is faster and more accurate than both its one- and two-stage precursors.
Compared to two-stage anchor-based detectors~\cite{cai2018cascade}, our first stage is more accurate and allows the detector to use fewer proposals in RoI heads (256 vs.\ 1K), making the detector both more accurate and faster overall.
Compared to single-stage detectors, our first stage uses a leaner head design and only has one output class for dense image-level prediction.
The speedup due to the drastic reduction in the number of classes more than makes up for the additional costs of the second stage.
Our second stage makes full use of years of progress in two-stage detection~\cite{cai2018cascade,chen2019hybrid} and yields a significant increase in detection accuracy over one-stage baselines.
It also easily scales to large-vocabulary detection.

Experiments on COCO~\cite{lin2014coco}, LVIS~\cite{gupta2019lvis}, and Objects365~\cite{shao2019objects365} demonstrate that our probabilistic two-stage framework boosts the accuracy of a strong CascadeRCNN model by 1-3 mAP, while also improving its speed.
Using a standard ResNeXt-101-DCN backbone with a CenterNet~\cite{zhou2019objects} first stage, our detector achieves 50.2 mAP on COCO test-dev.
With a strong Res2Net-101-DCN-BiFPN~\cite{gao2019res2net,tan2020efficientdet} backbone and self-training~\cite{zoph2020rethinking}, it achieves 56.4 mAP with single-scale testing, outperforming all published results.
Using a small DLA-BiFPN backbone and lower input resolution, we achieve 49.2 mAP on COCO at 33 fps on a Titan Xp, outperforming the popular YOLOv4 model (43.5 mAP at 33 fps) on the same hardware.
Code and models are release at \url{https://github.com/xingyizhou/CenterNet2}.

\section{Related Work}

\noindent \textbf{One-stage detectors} jointly predict an output class and location of objects densely throughout the image.
RetinaNet~\cite{lin2018focal} classifies a set of predefined sliding anchor boxes and handles the foreground-background imbalance by reweighting losses for each output.
FCOS~\cite{tian2019fcos} and CenterNet~\cite{zhou2019objects} eliminate the need of multiple anchors per pixel and classify foreground/background by location.
ATSS~\cite{zhang2020bridging} and PAA~\cite{paa-eccv2020} further improve FCOS by changing the definition of foreground and background.
GFL~\cite{li2020generalized} and Autoassign~\cite{zhu2020autoassign} change the hard foreground-background assignment to a weighted soft assignment.
AlignDet~\cite{chen2019revisiting} uses a deformable convolution layer before the output to gather richer features for classification and regression.
RepPoint~\cite{yang2019reppoints} and DenseRepPoint~\cite{yang2019dense} encode bounding boxes as the outline of a set of points and use the features of the point set for classification.
BorderDet~\cite{qiu2020borderdet} pools features along the bounding box for better localization.
Most one-stage detectors have a sound probabilistic interpretation.

While one-stage detectors have achieved competitive performance~\cite{zhang2020bridging,paa-eccv2020,zhang2019freeanchor,li2020generalized,zhu2020autoassign}, they usually rely on heavier separate classification and regression branches than two-stage models.
In fact, they are no longer faster than their two-stage counterparts if the vocabulary (i.e., the set of object classes) is large (as in the LVIS or Objects365 datasets).
Also, one-stage detectors only use the local feature of the positive cell for regression and classification, which is sometimes misaligned with the object~\cite{chen2019revisiting,song2020revisiting}.

Our probabilistic two-stage framework retains the probabilistic interpretation of one-stage detectors, but factorizes the probability distribution over multiple stages, improving both accuracy and speed.

\noindent \textbf{Two-stage detectors}
first use a region proposal network (RPN) to generate coarse object proposals, and then use a dedicated per-region head to classify and refine them.
FasterRCNN~\cite{ren2015faster,he2017mask} uses two fully-connected layers as the RoI heads.
CascadeRCNN~\cite{cai2018cascade} uses three cascaded stages of FasterRCNN, each with a different positive threshold so that the later stages focus more on localization accuracy.
HTC~\cite{chen2019hybrid} utilizes additional instance and semantic segmentation annotations to enhance the inter-stage feature flow of CascadeRCNN.
TSD~\cite{song2020revisiting} decouples the classification and localization branches for each RoI.

Two-stage detectors are still more accurate in many settings~\cite{gupta2019lvis,sun2019scalability,kuznetsova2018open}.
Currently, all two-stage detectors use a relatively weak RPN that maximizes the recall of the top 1K proposals, and does not utilize the proposal score at test time.
The large number of proposals slows the system down, and the recall-based proposal network does not directly offer the same clear probabilistic interpretation as one-stage detectors.
Our framework addresses this, and integrates a strong class-agnostic single-stage object detector with later classification stages.
Our first stage uses fewer, but higher quality, regions, yielding both faster inference and higher accuracy.

\noindent \textbf{Other detectors.}
A family of object detectors identify objects via points in the image.
CornerNet~\cite{Law_2018_ECCV} detects the top-left and bottom-right corners and groups them using an embedding feature.
ExtremeNet~\cite{zhou2019bottomup} detects four extreme points and groups them using an additional center point.
\citet{duan2019centernet} detect the center point and use it to improve corner grouping.
Corner Proposal Net~\cite{duan2020corner} uses pairwise corner groupings as region proposals. CenterNet~\cite{zhou2019objects} detects the center point and regresses the bounding box parameters from it.

\begin{figure*}[t]
\centering
\begin{subfigure}[t]{0.19\linewidth}
\includegraphics[width=\linewidth,page=4]{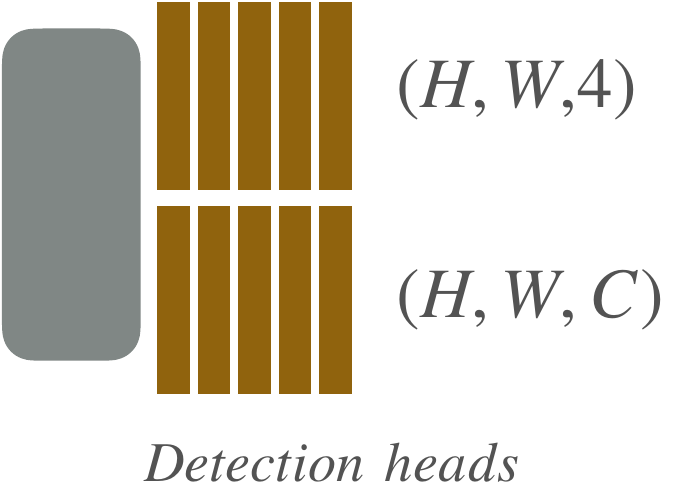}
\caption{one-stage detector}
\end{subfigure}\hfill
\begin{subfigure}[t]{0.33\linewidth}
\includegraphics[width=\linewidth,page=5]{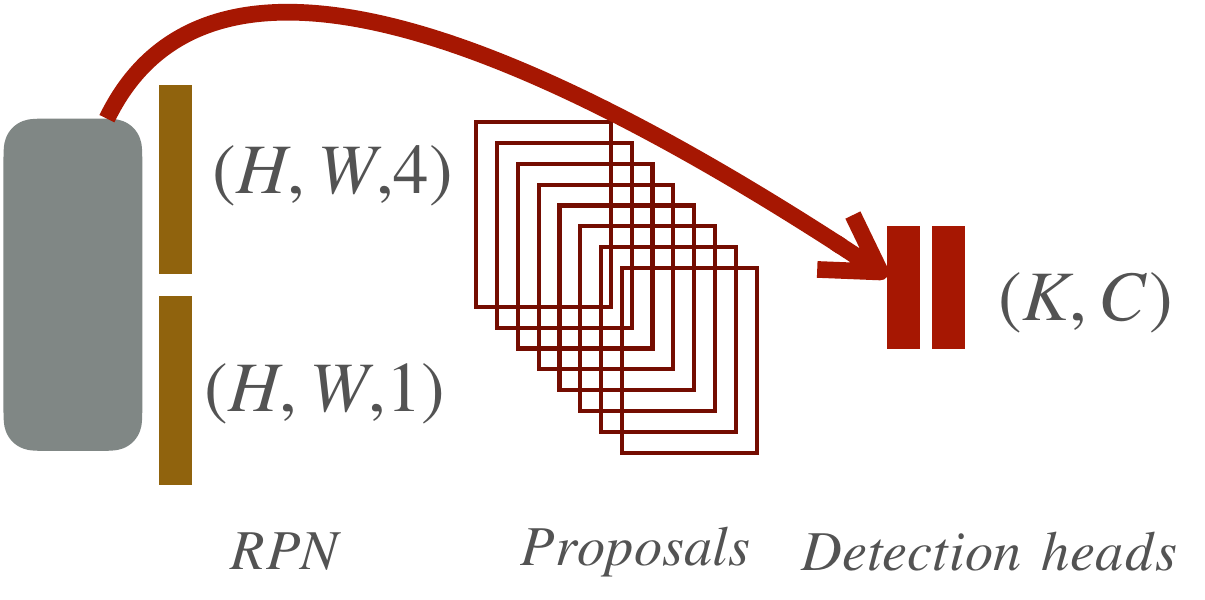}
\caption{two-stage detector}
\end{subfigure}\hfill
\begin{subfigure}[t]{0.33\linewidth}
\includegraphics[width=\linewidth,page=6]{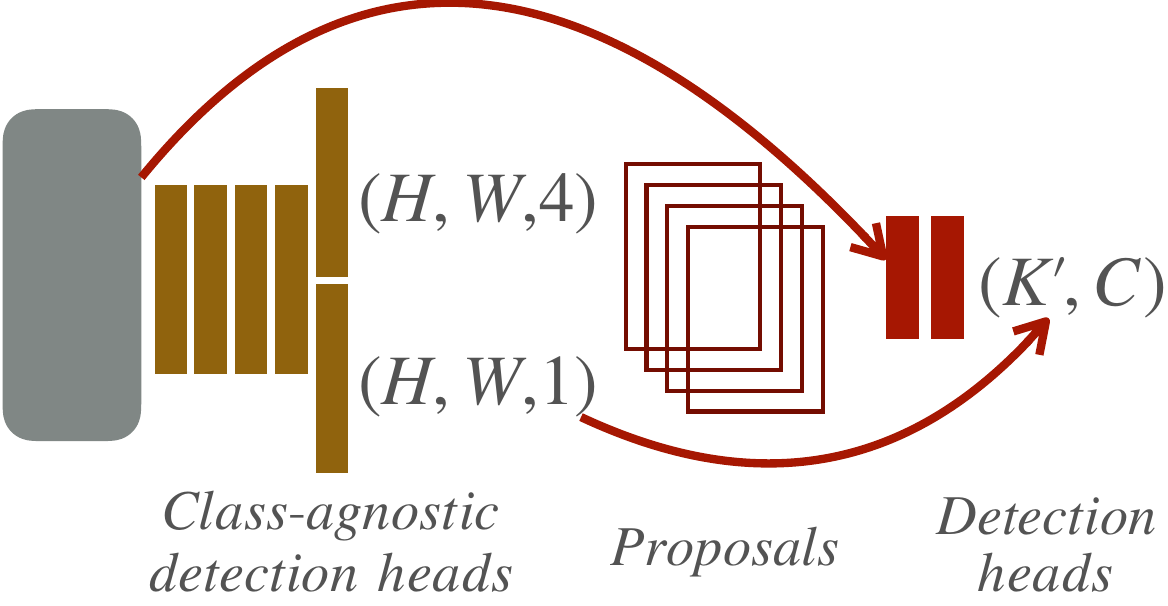}
\caption{Probabilistic two-stage detector}
\end{subfigure}%
\vspace{-3mm}
   \caption{
  Illustration of the structural differences between existing one-stage and two-stage detectors and our probabilistic two-stage framework. (a) A typical one-stage detector applies separate heavy classification and regression heads and produces a dense classification map. (b) A typical two-stage detector uses a light proposal network and extracts many ($K$) region features for classification. (c) Our probabilistic two-stage framework uses a one-stage detector with shared heads to produce region proposals and extracts a few ($K'$) regions for classification. The proposal score from the first stage is used in the second stage in a probabilistically sound framework. Typically, $K' < K \ll H \times W$.
   }
\label{fig:difference}
\vspace{-4mm}
\end{figure*}

DETR~\cite{carion2020end} and Deformable DETR~\cite{zhu2020deformable} remove the dense output in a detector, and instead use a Transformer~\cite{vaswani2017attention} that directly predicts a set of bounding boxes.

The major difference between point-based detectors, DETR, and conventional detectors lies in the network architecture.
Point-based detectors use a fully-convolutional network~\cite{newell2016stacked,yu2018deep}, usually with symmetric downsampling and upsampling layers, and produce a single feature map with a small stride (i.e., stride 4).
DETR-style detectors~\cite{carion2020end,zhu2020deformable} use a transformer as the decoder.
Conventional one- and two-stage detectors commonly use an image classification network augmented by lightweight upsampling layers, and produce multi-scale features (FPN)~\cite{lin2017feature}.

\section{Preliminaries}

An object detector aims to predict the location $b_i \in \mathbb{R}^4$ and class-specific likelihood score $s_i \in \mathbb{R}^{|\mathcal{C}|}$ for any object $i$ for a predefined set of classes $\mathcal{C}$.
The object location $b_i$ is most often described by two corners of an axis-aligned bounding box~\cite{ren2015faster,carion2020end} or through an equivalent center+size representation~\cite{tian2019fcos,zhou2019objects,zhu2020deformable}.
The main difference between object detectors lies in their representation of the class likelihood, reflected in their architectures.

\textbf{One-stage detectors}~\cite{redmon2018yolov3,lin2018focal,tian2019fcos,zhou2019objects} jointly predict the object location and likelihood score in a single network.
Let $L_{i,c}=1$ indicate a positive detection for object candidate $i$ and class $c$, and let $L_{i,c}=0$ indicate background.
Most one-stage detectors~\cite{lin2018focal,tian2019fcos,zhou2019objects} then parametrize the class likelihood as a Bernoulli distribution using an independent sigmoid per class: $s_i(c) = P(L_{i,c}=1) = \sigma(w_c^\top \vec f_i)$, where $f_i \in \mathbb{R}^C$ is a feature produced by the backbone and $w_c$ is a class-specific weight vector.
During training, this probabilistic interpretation allows one-stage detectors to simply maximize the log-likelihood $\log(P(L_{i,c}))$ or the focal loss~\cite{lin2018focal} of ground-truth annotations.
One-stage detectors differ from each other in the definition of positive $\hat L_{i,c}=1$ and negative $\hat L_{i,c}=0$ samples.
Some use anchor overlap~\cite{lin2018focal,zhang2020bridging,paa-eccv2020}, others use locations~\cite{tian2019fcos}.
However, all optimize log-likelihood and use the class probability to score boxes.
All directly regress to bounding box coordinates.

\textbf{Two-stage detectors}~\cite{ren2015faster,cai2018cascade} first extract potential object locations, called object proposals, using an objectness measure $P(O_i)$.
They then extract features for each potential object, classify them into $\mathcal{C}$ classes or background ${P(C_i | O_i=1)}$ with $C_i \in \mathcal{C} \cup \{bg\}$, and refine the object location.
Each stage is supervised independently.
In the first stage, a Region Proposal Network (RPN) learns to classify annotated objects $b_i$ as foreground and other boxes as background.
This is commonly done through a binary classifier trained with a log-likelihood objective.
However, an RPN defines background regions very conservatively.
Any prediction that overlaps an annotated object $30\%$ or more may be considered foreground.
This label definition favors recall over precision and accurate likelihood estimation.
Many partial objects receive a large proposal score.
In the second stage, a softmax classifier learns to classify each proposal into one of the foreground classes or background.
The classifier uses a log-likelihood objective, with foreground labels consisting of annotated objects and background labels coming from high-scoring first-stage proposals without annotated objects close-by.
During training, this categorical distribution is implicitly conditioned on positive detections of the first stage, as it is only trained and evaluated on them.
Both the first and second stage have a probabilistic interpretation, and 
under their positive and negative definition 
estimate the log-likelihood of objects or classes respectively.
However, the entire detector does not.
It combines multiple heuristics and sampling strategies to independently train the first and second stages~\cite{cai2018cascade,ren2015faster}.
The final output comprises boxes with classification scores $s_i(c) = P(C_i|O_i=1)$ of the second stage alone.

Next, we develop a simple probabilistic interpretation of two-stage detectors that considers the two stages as part of a single class-likelihood estimate.
We show how this affects the design of the first stage, and how to train the two stages efficiently.

\section{A probabilistic interpretation of two-stage detection}
For each image, our goal is to produce a set of $K$ detections as bounding boxes $b_1, \ldots, b_K$ with an associated class distribution $s_k(c) = P(C_k=c)$ for classes $c \in \mathcal{C} \cup \{bg\}$ or background to each object $k$.
In this work, we keep the bounding-box regression unchanged and only focus on the class distribution.
A two-stage detector factorizes this distribution into two parts: A class-agnostic object likelihood $P(O_k)$ (first stage) and a conditional categorical classification $P(C_k| O_k)$ (second stage).
Here $O_k=1$ indicates a positive detection in the first stage, while $O_k=0$ corresponds to background.
Any negative first-stage detection $O_k=0$ leads to a background $C_k=bg$ classification: $P(C_k=bg | O_k=0)=1$.
In a multi-stage detector~\cite{cai2018cascade}, the classification is done by an ensemble of multiple cascaded stages, while two-stage detectors use a single classifier~\cite{ren2015faster}.
The joint class distribution of the two-stage model then is
\begin{equation}
P(C_k) = \sum_o P(C_k|O_k=o)P(O_k=o)\lbleq{prob_det}.
\end{equation}

\paragraph{Training objective.} We train our detectors using maximum likelihood estimation.
For annotated objects, we maximize
\begin{equation}
 \log P(C_k) = \log P(C_k|O_k\!=\!1) + \log P(O_k\!=\!1)\lbleq{ll_pos},
\end{equation}
which reduces to independent maximum-likelihood objectives for the first and second stage respectively.

For the background class, the maximum-likelihood objective does not factorize:
\begin{equation*}
  \log P(bg) = \log \left(P(bg|O_k\!=\!1) P(O_k\!=\!1) + P(O_k\!=\!0)\right).
\end{equation*}
This objective ties the first- and second-stage probability estimates in their loss and gradient computation.
An exact evaluation requires a dense evaluation of the second stage for all first-stage outputs, which would slow down training prohibitively.
We instead derive two lower bounds to the objective, which we jointly optimize.
The first lower bound uses Jensen's inequality $\log\left(\alpha x_1+(1-\alpha)x_2\right)\ge \alpha \log(x_{1})+(1-\alpha)\log(x_{2})$ with $\alpha = P(O_k=1)$, $x_1 = P(bg|O_k\!=\!1)$, and $x_2=1$:
\begin{equation}
  \log P(bg) \ge P(O_k\!=\!1) \log \left(P(bg|O_k\!=\!1)\right) \lbleq{bound_sample}.
\end{equation}
This lower bound maximizes the log-likelihood of background of the second stage for any high-scoring object in the first stage.
It is tight for $P(O_k=1)\to0$ or $P(bg|O_k\!=\!1)\to 1$, but can be arbitrarily loose for $P(O_k=1)>0$ and $P(bg|O_k\!=\!1) \to 0$.
Our second bound involves just the first-stage objective:
\begin{equation}
  \log P(bg) \ge \log \left(P(O_k\!=\!0)\right) \lbleq{bound_independent}.\\
\end{equation}
It uses $P(bg|O_k\!=\!1) P(O_k\!=\!1) \ge 0$ with the monotonicity of the $\log$.
This bound is tight for $P(bg|O_k\!=\!1) \to 0$.
Ideally, the tightest bound is obtained by using the maximum of \shortrefeq{bound_sample} and \shortrefeq{bound_independent}.
This lower bound is within $\le \log 2$ of the actual objective, as shown in the supplementary material.
In practice however, we found optimizing both bounds jointly to work better.

With lower bound \shortrefeq{bound_independent} and the positive objective \shortrefeq{ll_pos}, first-stage training reduces to a maximum-likelihood estimate with positive labels at annotated objects and negative labels for all other locations.
It is equivalent to training a binary one-stage detector, or an RPN with a strict negative definition that encourages likelihood estimation and not recall.

\paragraph{Detector design.}
The key difference between our formulation and standard two-stage detectors lies in the use of the class-agnostic detection $P(O_k)$ in the detection score \shortrefeq{prob_det}.
In our probabilistic formation, the classification score is multiplied by the class-agnostic detection score.
This requires a strong first stage detector that not only maximizes the proposal recall~\cite{ren2015faster,uijlings2013selective}, but also predicts a reliable object likelihood for each proposal.
In our experiments, we use strong one-stage detectors to estimate this log-likelihood, as described in the next section.

\section{Building a probabilistic two-stage detector}
\lblsec{designing}

The core component of a probabilistic two-stage detector is a strong first stage.
This first stage needs to predict an accurate object likelihood that informs the overall detection score, rather than maximizing the object coverage.
We experiment with four different first-stage designs based on popular one-stage detectors.
For each, we highlight the design choices needed to convert them from a single-stage detector to a first stage in a probabilistic two-stage detector.

\textbf{RetinaNet}~\cite{lin2018focal} closely resembles the RPN of traditional two-stage detectors with three critical differences: a heavier head design (4 layers vs.\ 1 layer in RPN), a stricter positive and negative anchor definition, and the focal loss.
Each of these components increases RetinaNet's ability to produce calibrated one-stage detection likelihoods.
We use all of these in our first-stage design.
RetinaNet by default uses two separate heads for bounding box regression and classification.
In our first-stage design, we found it sufficient to have a single shared head for both tasks, as object-or-not classification is easier and requires less network capacity.
This speeds up inference.

\textbf{CenterNet}~\cite{zhou2019objects} finds objects as keypoints located at their center, then regresses to box parameters.
The original CenterNet operates at a single scale,
whereas conventional two-stage detectors use a feature pyramid (FPN)~\cite{lin2017feature}.
We upgrade CenterNet to multiple scales using an FPN.
Specifically, we use the RetinaNet-style ResNet-FPN as the backbone~\cite{lin2018focal}, with output feature maps from stride 8 to 128 (i.e., P$3$-P$7$). 
We apply a 4-layer classification branch and regression branch~\cite{tian2019fcos} to all FPN levels to produce a detection heatmap and bounding box regression map.
During training, we assign ground-truth center annotations to specific FPN levels based on the object size, within a fixed assignment range~\cite{tian2019fcos}. 
Inspired by GFL~\cite{li2020generalized}, we add locations in the $3\times3$ neighborhood of the center that already produce high-quality bounding boxes (i.e., with a regression loss $<0.2$) as positives.
We use the distance to boundaries as the bounding box representation~\cite{tian2019fcos}, and use the gIoU loss for bounding box regression~\cite{rezatofighi2019generalized}.
We evaluate both one-stage and probabilistic two-stage versions of this architecture.
We refer to the improved CenterNet as CenterNet*.

\textbf{ATSS}~\cite{zhang2020bridging} models the class likelihood of a one-stage detector with an adaptive IoU threshold for each object, and uses centerness~\cite{tian2019fcos} to calibrate the score.
In a probabilistic two-stage baseline, we use ATSS~\cite{zhang2020bridging} as is, and multiply the centerness and the foreground classification score for each proposal.
We again merge the classification and regression heads for a slight speedup.

\textbf{GFL}~\cite{li2020generalized} uses regression quality to guide the object likelihood training.
In a probabilistic two-stage baseline, we remove the integration-based regression and only use the distance-based regression~\cite{tian2019fcos} for consistency, and again merge the two heads.

The above one-stage architectures infer $P(O_k)$.
For each, we combine them with the second stage that infers $P(C_k |O_k)$.
We experiment with two basic second-stage designs: FasterRCNN~\cite{ren2015faster} and CascadeRCNN~\cite{cai2018cascade}.

\paragraph{Hyperparameters.}
A two-stage detector~\cite{ren2015faster} typically uses FPN levels P2-P6 (stride 4 to stride 64), while most one-stage detectors use FPN levels P3-P7 (stride 8 to stride 128).
To make it compatible, we use levels P3-P7 for both one- and two-stage detectors.
This modification slightly improves the baselines.
Following \citet{wang2019region}, we increase the positive IoU threshold in the second stage from $0.5$ to $0.6$ for Faster RCNN (and $0.6, 0.7, 0.8$ for CascadeRCNN)
to compensate for the IoU distribution change in the second stage.
We use a maximum of 256 proposal boxes in the second stage for probabilistic two-stage detectors, and use the default 1K boxes for RPN-based models unless stated otherwise.
We also increase the NMS threshold from $0.5$ to $0.7$ for our probabilistic detectors as we use fewer proposals.
These hyperparameter-changes is necessary for probabilistic detectors, but we found they do not improve the RPN-based detector in our experiments.

We implement our method based on detectron2~\cite{wu2019detectron2}.
Our default model follows the standard setting in detectron2~\cite{wu2019detectron2}.
Specifically, we train the network with the SGD optimizer for 90K iterations (1x schedule).
The base learning rate is $0.02$ for two-stage detectors and $0.01$ for one-stage detectors, 
and is dropped by 10x at iterations 60K and 80K.
We use multi-scale training with the short edge in the range [640,800] and the long edge up to 1333.
During training, we set the first-stage loss weight to $0.5$ as one-stage detectors are typically trained with learning rate $0.01$.
During testing, we use a fixed short edge at 800 and long edge up to 1333.

We instantiate our probabilistic two-stage framework on four different backbones.
We use a default ResNet-50~\cite{he2016deep} model for most ablations and comparisons among design choices, and then compare to state-of-the-art methods using the same large ResNeXt-32x8d-101-DCN~\cite{xie2017aggregated} backbone, and use a lightweight DLA~\cite{yu2018deep} backbone for a real-time model.
We also integrate the most recent advances~\cite{zoph2020rethinking,tan2020efficientdet,gao2019res2net} and design an extra-large backbone for the high-accuracy regime.
Further details about each backbone are in the supplement.

{
\begin{table}[t]
\centering
\begin{tabular}{l@{\ \ }c@{\ \ \ \ }c@{\ \ \ \ }c}
\toprule
& mAP & $T_{first}$ & $T_{tot}$ \\
\midrule
FasterRCNN-RPN (original) & 37.9 & 46ms & 55ms \\
CascadeRCNN-RPN (original) & 41.6 & 48ms & 78ms \\
\midrule
RetinaNet~\cite{lin2018focal} & 37.4 & 82ms & 82ms \\
\rowcolor{lightgray}
FasterRCNN-RetinaNet & 40.4 & 60ms & 63ms \\
\rowcolor{lightgray}
CascadeRCNN-RetinaNet & 42.6 & 61ms & 69ms \\
\midrule
GFL~\cite{li2020generalized} & 40.2 & 51ms & 51ms\\
\rowcolor{lightgray}
FasterRCNN-GFL & 41.7 & 46ms & 50ms \\
\rowcolor{lightgray}
CascadeRCNN-GFL & 42.7 & 46ms & 57ms \\
\midrule
ATSS~\cite{zhang2020bridging} & 39.7 & 56ms & 56ms\\
\rowcolor{lightgray}
FasterRCNN-ATSS & 41.5 & 47ms & 50ms \\
\rowcolor{lightgray}
CascadeRCNN-ATSS & 42.7 & 47ms & 57ms \\
\midrule
CenterNet* & 40.2 & 51ms & 51ms\\
\rowcolor{lightgray}
FasterRCNN-CenterNet & 41.5 & 46ms & 50ms \\
\rowcolor{lightgray}
CascadeRCNN-CenterNet & \textbf{42.9} & 47ms & 57ms\\
\bottomrule
\end{tabular}
\vspace{-2mm}
\caption{Performance and runtime of a number of two-stage detectors, one-stage detectors, and corresponding \ctext[RGB]{238,238,236}{probabilistic two-stage detectors} (our approach). Results on COCO validation.
Top block: two-stage FasterRCNN and CascadeRCNN detectors.
Other blocks: Four one-stage detectors, each with two corresponding \ctext[RGB]{238,238,236}{probabilistic two-stage detectors}, one based on FasterRCNN and one based on CascadeRCNN.
For each detector, we list its first-stage runtime ($T_{first}$) and total runtime ($T_{tot}$).
All results are reported using standard Res50-1x with multi-scale training.
}
\label{table:rpn}
\vspace{-5mm}
\end{table}
}

\section{Results}

We evaluate our framework on three large detection datasets: COCO~\cite{lin2014coco}, LVIS~\cite{gupta2019lvis}, and Objects365~\cite{gao2019objects365}.
Details of each dataset can be found in the supplement.
We use COCO to perform ablation studies and comparisons to the state of the art.
We use LVIS and Objects365 to test the generality of our framework, particularly in the large-vocabulary regime.
In all datasets, we report the standard mAP.
Runtimes are reported on a Titan Xp GPU with PyTorch 1.4.0 and CUDA 10.1.

Table~\ref{table:rpn} compares one- and two-stage detectors to corresponding probabilistic two-stage detectors designed via our framework.
The first block of the table shows the performance of the original reference two-stage detectors, FasterRCNN and CascadeRCNN.
The following blocks show the performance of four one-stage detectors (discussed in Section~\ref{sec:designing}) and the corresponding probabilistic two-stage detectors, obtained when using the respective one-stage detector as the first stage in a probabilistic two-stage framework. For each one-stage detector, we show two versions of probabilistic two-stage models, one based on FasterRCNN and one based on CascadeRCNN.

\begin{table}[t]
\centering
\begin{tabular}{@{}l@{\ \ }c@{\ \ }c@{\ \ }c@{\ \ }c@{}}
\toprule
& Backbone & Epochs & mAP & Runtime \\
\midrule
FCOS-RT & DLA-BiFPN-P3 & 48 & 42.1 & 21ms \\
\name & DLA-BiFPN-P3 & 48 & 43.7 & 25ms \\
\midrule
CenterNet & DLA & 230 & 37.6 & 18ms \\
YOLOV4 & CSPDarknet-53 & 300 & 43.5 & 30ms \\
EfficientDet & EfficientNet-B2 & 500 & 43.5 & \textcolor{lightgray}{23ms}* \\
EfficientDet & EfficientNet-B3 & 500 & 46.8 & \textcolor{lightgray}{37ms}* \\
\name & DLA-BiFPN-P3 & 288 & 45.6 & 25ms \\
\midrule
\name & DLA-BiFPN-P5 & 288 & 49.2 & 30ms \\
\bottomrule
\end{tabular}
\vspace{-2mm}
\caption{Performance of real-time object detectors on COCO validation. Top: we compare \name to realtime-FCOS under exactly the same setting. Bottom: we compare to detectors with different backbones and training schedules. *The runtime of EfficientDet is taken from the original paper~\cite{tan2020efficientdet} as the official model is not available. Other runtimes are measured on the same machine.}
\label{table:realtime}
\vspace{-5mm}
\end{table}

\begin{table*}[t]
\centering
\begin{tabular}{@{}l@{\ \ \ \ }c@{\ \ \ \ }c@{\ \ \ \ }c@{\ \ \ \ }c@{\ \ \ \ }c@{\ \ \ \ }c@{\ \ \ \ }c@{\ \ \ \ }c@{\ \ \ \ }c@{\ \ \ \ }c@{}}
\toprule
& Backbone & $AP$ & $AP_{50}$ & $AP_{75}$ & $AP_{S}$ & $AP_{M}$ & $AP_{L}$  \\
\midrule
        CornerNet~\cite{Law_2018_ECCV} & Hourglass-104 & 40.6 & 56.4 & 43.2 & 19.1 & 42.8 & 54.3 \\
        CenterNet~\cite{zhou2019objects} & Hourglass-104 & 42.1 & 61.1& 45.9 & 24.1 & 45.5 & 52.8\\
        Duan et al.~\cite{duan2019centernet} & Hourglass-104 &  44.9 & 62.4 & 48.1 & 25.6 & 47.4 & 57.4 \\
        RepPoint~\cite{yang2019reppoints} & ResNet101-DCN &45.0 & 66.1 & 49.0 & 26.6 & 48.6 & 57.5 \\
        MAL~\cite{ke2020multiple} & ResNeXt-101 &  45.9 & 65.4 & 49.7 & 27.8 & 49.1 & 57.8 \\
        FreeAnchor~\cite{zhang2019freeanchor} & ResNeXt-101 & 46.0 & 65.6 & 49.8 & 27.8 & 49.5 & 57.7 \\
        CentripetalNet~\cite{Dong_2020_CVPR} & Hourglass-104 & 46.1 & 63.1 & 49.7 & 25.3 & 48.7 & 59.2 \\
        FCOS~\cite{tian2019fcos} & ResNeXt-101-DCN & 46.6 & 65.9 & 50.8 & 28.6 & 49.1 & 58.6 \\
        TridentNet~\cite{li2019scale} & ResNet-101-DCN & 46.8 & 67.6 & 51.5 & 28.0 & 51.2 & 60.5 \\
        CPN~\cite{duan2020corner} & Hourglass-104 & 47.0 & 65.0 & 51.0 & 26.5 & 50.2 & 60.7\\
        SAPD~\cite{zhu2019soft} & ResNeXt-101-DCN & 47.4 & 67.4 & 51.1 & 28.1 & 50.3 & 61.5\\
        ATSS~\cite{zhang2020bridging} & ResNeXt-101-DCN & 47.7 & 66.6 & 52.1 & 29.3 & 50.8 & 59.7 \\
        BorderDet~\cite{yang2019reppoints} & ResNeXt-101-DCN & 48.0 & 67.1 & 52.1 & 29.4 & 50.7 & 60.5 \\
        GFL~\cite{li2020generalized} & ResNeXt-101-DCN &  48.2 & 67.4 & 52.6 & 29.2 & 51.7 & 60.2 \\
        PAA~\cite{paa-eccv2020} & ResNeXt-101-DCN & 49.0 & 67.8 & 53.3 & 30.2 & 52.8 & 62.2 \\
        TSD~\cite{song2020revisiting} & ResNeXt-101-DCN & 49.4 & 69.6 & 54.4 & \textbf{32.7} & 52.5 & 61.0 \\
        RepPointv2~\cite{yang2019reppoints} & ResNeXt-101-DCN & 49.4 & 68.9 & 53.4 & 30.3 & 52.1 & 62.3 \\
        AutoAssign~\cite{zhu2020autoassign} & ResNeXt-101-DCN & 49.5 & 68.7 & 54.0 & 29.9 & 52.6 & 62.0 \\
        Deformable DETR~\cite{zhu2020deformable} & ResNeXt-101-DCN & 50.1 & \textbf{69.7} & 54.6 & 30.6 & 52.8 & \textbf{65.6} \\
        CascadeRCNN~\cite{cai2018cascade} & ResNeXt-101-DCN & 48.8 & 67.7 & 52.9 & 29.7 & 51.8 & 61.8 \\
        CenterNet* & ResNeXt-101-DCN & 49.1 & 67.8 & 53.3 & 30.2 & 52.4 & 62.0\\
        \name (ours) & ResNeXt-101-DCN & \textbf{50.2} & 68.0 & \textbf{55.0} & 31.2 & \textbf{53.5} & 63.6\\
        \midrule
        CRCNN-ResNeSt~\cite{zhang2020resnest} & ResNeSt-200 & 49.1 & 67.8 & 53.2 & 31.6 & 52.6 & 62.8 \\
        GFLV2~\cite{li2020generalizedv2} & Res2Net-101-DCN & 50.6 & 69.0 & 55.3 & 31.3 & 54.3 & 63.5 \\
        DetectRS~\cite{qiao2020detectors} & ResNeXt-101-DCN-RFP & 53.3 & 71.6 & 58.5 & 33.9 & 56.5 & 66.9 \\
        EfficientDet-D7x~\cite{tan2020efficientdet} & EfficientNet-D7x-BiFPN & 55.1 & {73.4} & 59.9 & - & - & - \\
        ScaledYOLOv4~\cite{wang2020scaled} & CSPDarkNet-P7 & 55.4 & { 73.3} & 60.7 & {38.1} & {59.5} & 67.4 \\
        CenterNet2 (ours) & Res2Net-101-DCN-BiFPN & {\bf 56.4} & {\bf 74.0} & {\bf 61.6} & {\bf 38.7} & {\bf 59.7} & {\bf 68.6} \\
\bottomrule
\end{tabular}
\normalsize
\vspace{-2mm}
\caption{Comparison to the state of the art on COCO test-dev. We list object detection accuracy with single-scale testing. We retrained our baselines, CascadeRCNN (ResNeXt-101-DCN) and CenterNet*, under comparable settings. Other results are taken from the original publications. Top: detectors with comparable backbones (ResNeXt-101-DCN) and training schedules (2x). Bottom: detectors with their best-fit backbones, input size, and schedules.}
\label{table:sota}
\vspace{-5mm}
\end{table*}

All probabilistic two-stage detectors outperform their one-stage and two-stage precursors.
Each probabilistic two-stage FasterRCNN model improves upon its one-stage precursor by 1 to 2 percentage points in mAP, and outperforms the original two-stage FasterRCNN by up to 3 percentage points in mAP.
More interestingly, each two-stage probabilistic FasterRCNN is \emph{faster} than its one-stage precursor due to the leaner head design.
A number of probabilistic two-stage FasterRCNN models are faster than the original two-stage FasterRCNN, due to more efficient FPN levels (P3-P7 vs.\ P2-P6) and because the probabilistic detectors 
use
fewer proposals (256 vs.\ 1K).
We observe similar trends with the CascadeRCNN models.

The CascadeRCNN-CenterNet design performs best among these probabilistic two-stage models. We thus adopt this basic structure in the following experiments and refer to it as CenterNet2 for brevity.

\paragraph{Real-time models.}
Table~\ref{table:realtime} compares our real-time model to other real-time detectors.
CenterNet2 outperforms realtime-FCOS~\cite{tian2020fcos} by $1.6$ mAP with the same backbone and training schedule, and is only 4 ms slower.
Using the same FCOS-based backbone with longer training schedules~\cite{tan2020efficientdet,bochkovskiy2020yolov4}, it improves upon the original CenterNet~\cite{zhou2019objects} by $7.7$ mAP, and comfortably outperforms the popular YOLOv4~\cite{bochkovskiy2020yolov4} and EfficientDet-B2~\cite{tan2020efficientdet} detectors with $45.6$ mAP at 40 fps.
Using a slightly different FPN structure and combining with self-training ~\cite{zoph2020rethinking}, CenterNet2 gets $49.2$ mAP at 33 fps.
While most existing real-time detectors are one-stage, 
here we show that two-stage detectors can be as fast as one-stage designs, while delivering higher accuracy.

\paragraph{State-of-the-art comparison.}
Table~\ref{table:sota} compares our large models to state-of-the-art detectors on COCO test-dev.
Using a ``standard'' large backbone ResNeXt101-DCN, \name achieves 50.2 mAP, outperforming all existing models with the same backbone, both one- and two-stage. 
Note that CenterNet2 outperforms the corresponding CascadeRCNN model with the same backbone by 1.4 percentage points in mAP. This again highlights the benefits of a probabilistic treatment of two-stage detection.

To push the state-of-the-art of object detection, we further switch to a stronger backbone Res2Net~\cite{gao2019res2net} with BiFPN~\cite{tan2020efficientdet}, a larger input resolution ($1280 \times 1280$ in training and $1560 \times 1560$ in testing) with heavy crop augmentation (ratio 0.1 to 2)~\cite{tan2020efficientdet}, and a longer schedule ($8\times$) with self-training ~\cite{zoph2020rethinking} on COCO unlabeled images.
Our final model achieves $56.4$ mAP with a single model, outperforming all published numbers in the literature. 
More details about the extra-large model can be found in the supplement.

{
\begin{table}[!t]
\centering
\begin{tabular}{c@{\ \ }c@{\ \ \ }c@{\ \ \ }c@{\ \ }c@{\ \ \ }ccc}
\toprule
P3-P7 & 256p. & 4 l. & loss & prob & mAP & $T_{first}$ & $T_{tot}$ \\
\midrule
 & & & & & 37.9 & 46ms & 55ms \\
\midrule
 \checkmark & & & & & 38.6 & 38ms & 45ms \\
 \checkmark & & & &  \checkmark & 38.5 & 38ms & 45ms \\
 \checkmark & \checkmark & & & & 38.3 & 38ms & 40ms \\
 \checkmark & & \checkmark & & & 38.9 & 60ms & 70ms \\
 \checkmark & \checkmark & \checkmark & & & 38.6 & 60ms & 63ms \\
\checkmark & \checkmark & \checkmark & \checkmark & & 39.1 & 60ms & 63ms \\
\midrule
\rowcolor{lightgray}
\checkmark & \checkmark & \checkmark & \checkmark & \checkmark & 40.4 & 60ms & 63ms \\
\bottomrule
\end{tabular}
\vspace{-2mm}
\caption{A detailed ablation between FasterRCNN-RPN (top) and a probabilistic two-stage FasterRCNN-RetinaNet (bottom). FasterRCNN-RetinaNet changes the FPN levels (P2-P6 to P3-P7), uses 256 instead of 1000 proposals, a 4-layer first-stage head, a stricter IoU threshold with focal loss (loss), and multiplies the first and second stage probabilities (prob).
All results are reported using standard Res50-1x with multi-scale training.
}
\label{table:rpn_vs_RetinaNet}
\vspace{-5mm}
\end{table}
}

\begin{table}[!t]
\centering
\begin{tabular}{@{}l@{\ \ \ }c@{}}
\toprule
& mAP \\
\midrule
CascadeRCNN-RPN (P3-P7) & 42.1 \\
CascadeRCNN-RPN w. prob. & 42.1 \\
CascadeRCNN-CenterNet & 42.1 \\
\rowcolor{lightgray}%
CascadeRCNN-CenterNet w. prob. (Ours) & 42.9 \\
\bottomrule
\end{tabular}
\vspace{-2mm}
\caption{
Ablation of our probabilistic modeling (w. prob.) of CascadeRCNN with the default RPN and CenterNet proposal.
}
\label{tab:hyperparameters}
\vspace{-6mm}
\end{table}

\begin{table}[!t]
\centering
\begin{tabular}{@{}l@{\ \ }c@{\ \ \ }c@{\ \ \ }c@{\ \ \ }c@{\ \ \ }c@{\ \ \ }c}
\toprule
 & \multicolumn{3}{c}{CascadeRCNN} & \multicolumn{3}{c}{CenterNet2} \\
\#prop. & mAP & AR & Runtime & mAP & AR & Runtime \\
\cmidrule(r){1-1}
\cmidrule(r){2-4}
\cmidrule(r){5-7}
\rowcolor{lightgray}%
1000 & 42.1 & 62.4 &  66ms & \cellcolor{white}43.0 & \cellcolor{white}70.8  & \cellcolor{white}75ms \\
512 & 41.9 & 60.4 & 56ms & 42.9 & 69.0 & 61ms \\
\rowcolor{lightgray}%
\cellcolor{white}256 & \cellcolor{white}41.6 & \cellcolor{white}57.4 & \cellcolor{white} 48ms & 42.9 &  66.6 & 57ms \\
128 & 40.8 & 53.7 & 45ms & 42.7 & 63.5 & 54ms \\
64 & 39.6 & 49.2 & 42ms & 42.1 & 59.7 & 52ms \\
\bottomrule
\end{tabular}
\vspace{-2mm}
\caption{Accuracy-runtime trade-off of using different numbers of proposals (\#prop.) on COCO validation. We show the overall mAP, the proposal recall (AR), and runtime for both the original CascadeRCNN and our probabilistic two-stage detector (CenterNet2). The results are reported with Res50-1x and multi-scale training. We highlight the default number of proposals in \colorbox{lightgray}{gray}.}
\label{table:num_proposals}
\vspace{-6mm}
\end{table}

{
\begin{figure*}[t]
\centering
     \begin{tabular}{c@{}c@{}c@{}c}
      \includegraphics[width=0.24\textwidth]{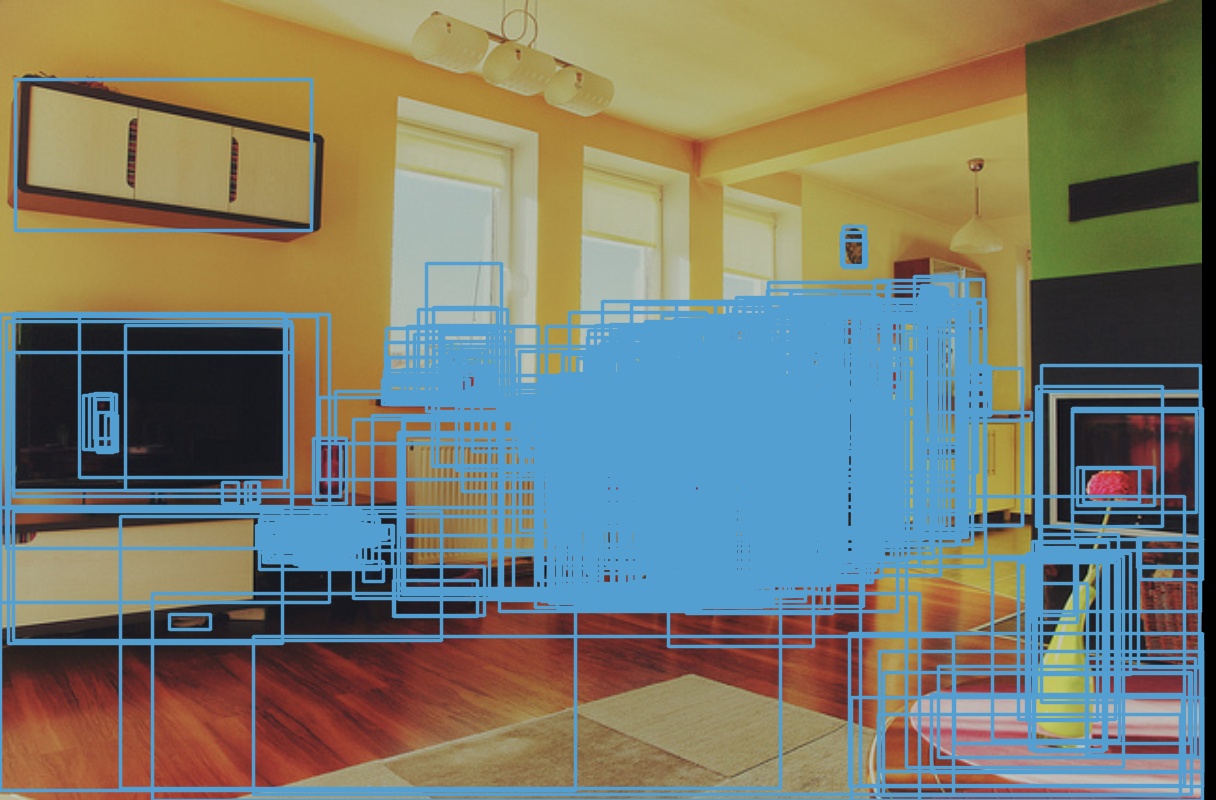}
      &\includegraphics[width=0.24\textwidth]{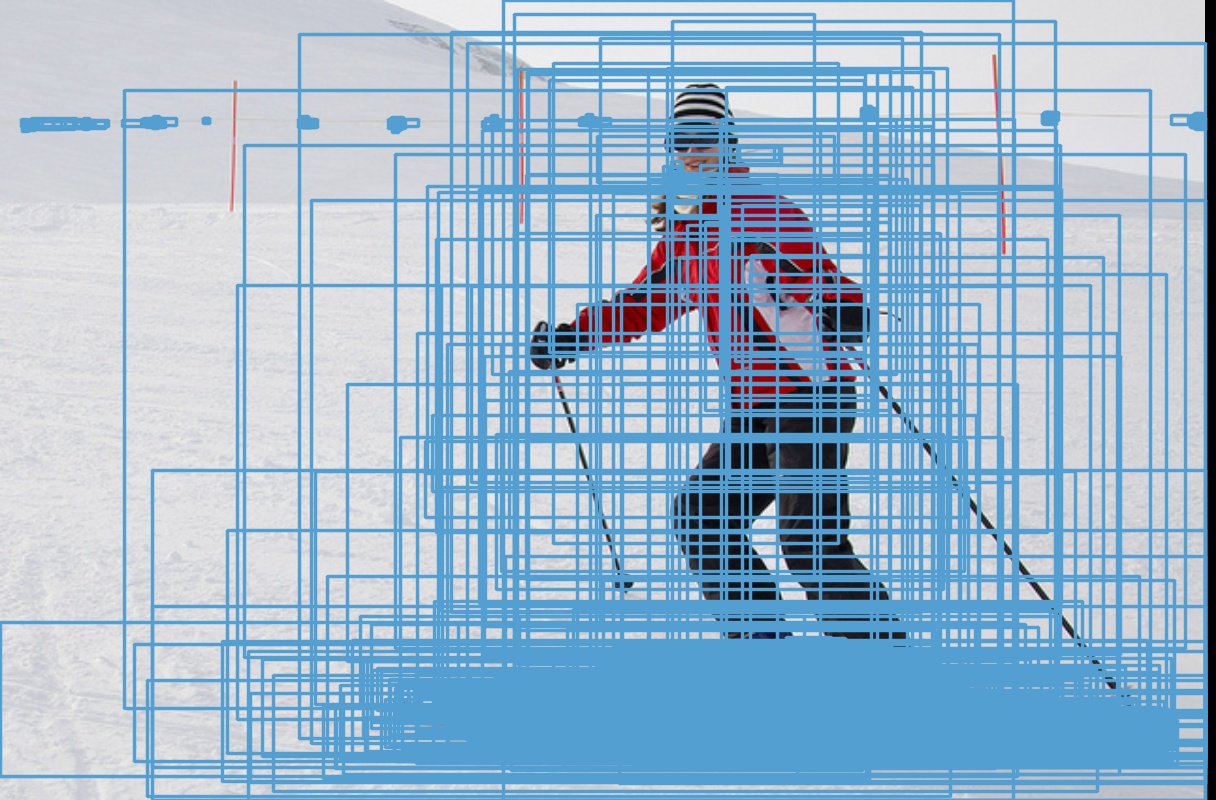}
      \includegraphics[width=0.24\textwidth]{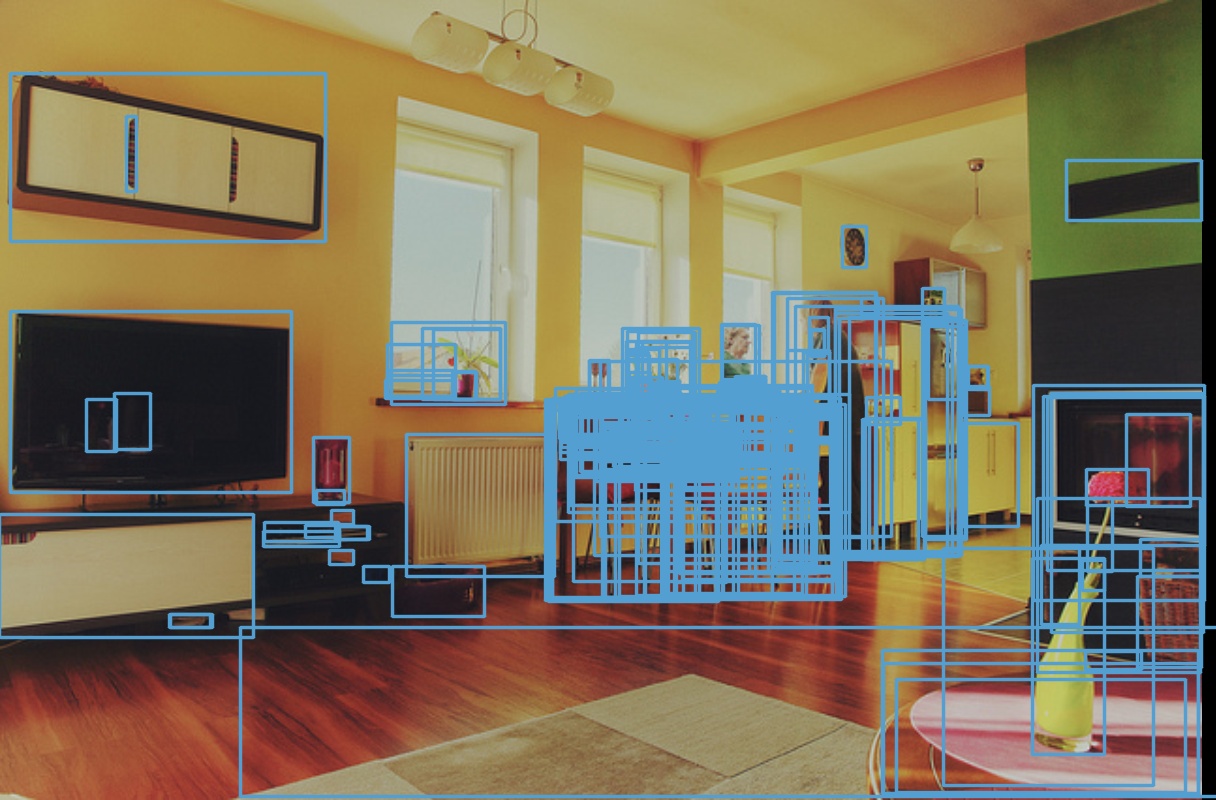}
      &\includegraphics[width=0.24\textwidth]{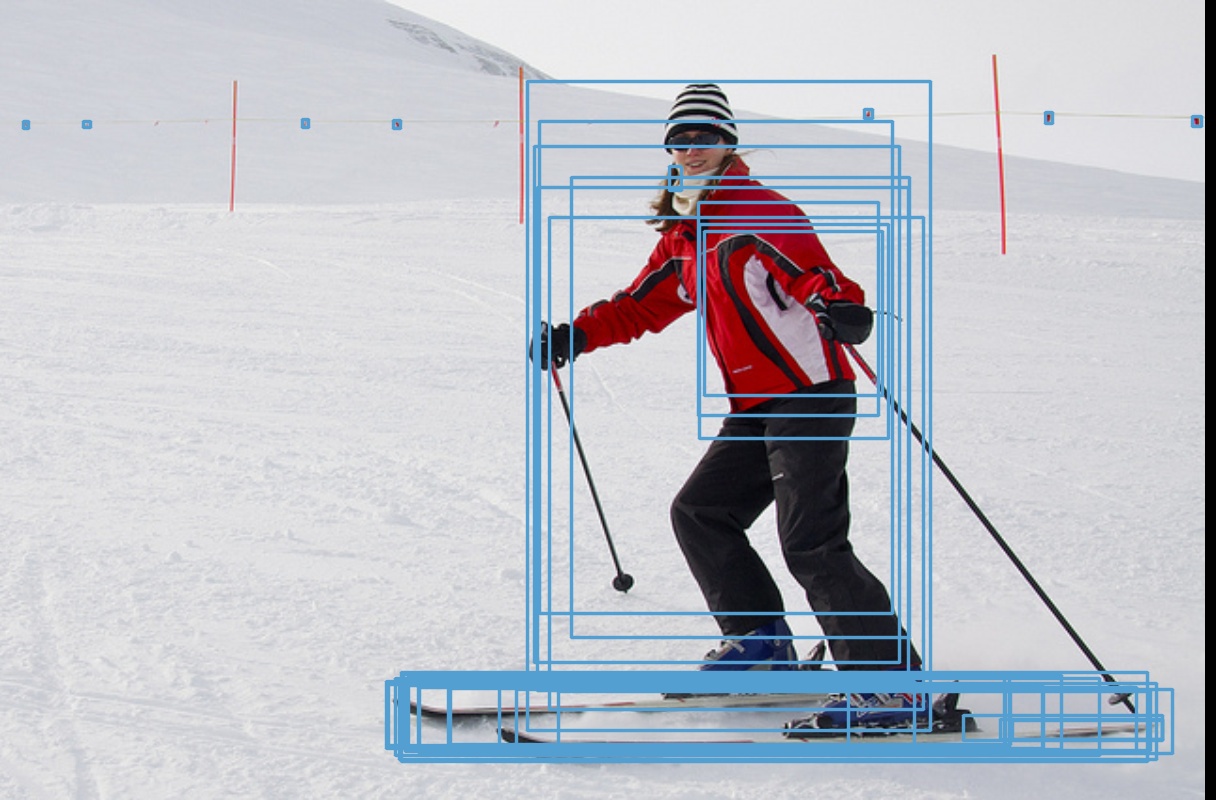}
      \end{tabular}
      \vspace{-3mm}
      \caption{Visualization of region proposals on COCO validation, contrasting CascadeRCNN and its probabilistic counterpart, CascadeRCNN-CenterNet (or CenterNet2). Left: region proposals from the first stage of CascadeRCNN (RPN). Right: region proposals from the first stage of CenterNet2. For clarity, we only show regions with score $>$0.3. }
      \label{fig:demo}
      \vspace{-2mm}
\end{figure*}
}

\subsection{Ablation studies}
\label{sec:ablation}
\paragraph{From FasterRCNN-RPN to FasterRCNN-RetinaNet.}
Table~\ref{table:rpn_vs_RetinaNet} shows the detailed road map from the default RPN-FasterRCNN to a probabilistic two-stage FasterRCNN with RetinaNet as the first stage.
First, switching to the RetinaNet-style FPN already gives a favorable improvement.
However, directly multiplying the first-stage probability here does not give an improvement, because the original RPN is weak and does not provide a proper likelihood.
Making the RPN stronger by adding layers makes it possible to use fewer proposals in the second stage, but does not improve accuracy.
Switching to the RetinaNet loss (a stricter IoU threshold and focal loss), the proposal quality is improved, yielding a 0.5 mAP improvement over the original RPN loss.
With the improved proposals, incorporating the first-stage score in our probabilistic framework significantly boosts accuracy to 40.4. 

Table~\ref{tab:hyperparameters} reports similar ablations on CascadeRCNN.
The observations are consistent: multiplying the first-stage probabilities with the original RPN does not improve accuracy, while using a strong one-stage detector can.
This suggests that both ingredients in our design are necessary: a stronger proposal network and incorporating the proposal score.

\paragraph{Trade-off in the number of proposals.}
Table~\ref{table:num_proposals} shows how the mAP, proposal average recall (AR), and runtime change when using a different numbers of proposals for the original RPN-based CascadeRCNN and CenterNet2.
Both CascadeRCNN and CenterNet2 get faster with fewer proposals.
However, the accuracy of the original CascadeRCNN drops steeply as the number of proposals decreases, while our detector performs well even with relatively few proposals.
For example, CascadeCRNN drops by $1.3$ mAP when using 128 instead of 1000 proposals, while CenterNet2 only loses $0.3$ mAP.
The average recall of 128 CenterNet2 proposals is higher than 1000 RPN ones.

\begin{table}[!t]
\centering
\begin{tabular}{@{}l@{}c@{\ }c@{\ }c@{\ }c@{\ }c@{}}
\toprule
& mAP & mAP$_{r}$ & mAP$_{c}$ & mAP$_{f}$ & Runtime \\
\midrule
GFL{~\cite{li2020generalized}} & 18.5 & 6.9 & 15.8 & 26.6 & 69ms \\
CenterNet* & 19.1 & 7.8 & 16.3 & 27.4 & 69ms \\
CascadeRCNN & 24.0  & 7.6 & 22.9  & 32.7 & 100ms\\
\name & \bf 26.7  & \bf 12.0  & \bf 25.4 & \bf 34.5 & \textbf{60ms}\\
\midrule
\name w. FedLoss & \bf 28.2  & \bf 18.8  & \bf 26.4 & 34.4 & \textbf{60ms}\\
\bottomrule
\end{tabular}
\normalsize
\vspace{-2mm}
\caption{Object detection results on LVIS v1 validation. The experiments are conducted with Res50-1x, multi-scale training, and repeat-factor sampling~\cite{gupta2019lvis}. 
}
\vspace{-5mm}
\label{table:lvis}
\end{table}

\begin{table}[!t]
\centering
\begin{tabular}{@{}l@{\ \ }c@{\ \ }c@{\ \ }c@{\ \ }c@{}}
\toprule
& mAP & mAP$_{50}$ & mAP$_{75}$ & Runtime \\
\midrule
GFL~\cite{li2020generalized} & 18.8 & 28.1 & 20.2 & 56ms \\
CenterNet* & 18.7 & 27.5 & 20.1 & \textbf{55ms} \\
CascadeRCNN & 21.7  & 31.7 & 23.4 & 67ms \\
\name & \textbf{22.6} & \textbf{31.6}  & \textbf{24.6} & 56ms \\
\bottomrule
\end{tabular}
\normalsize
\vspace{-2mm}
\caption{Object detection results on Objects365. The experiments are conducted with Res50-1x, multi-scale training, and class-aware sampling~\cite{shen2016relay}. 
}
\vspace{-6mm}
\label{table:objects365}
\end{table}

\subsection{Large vocabulary detection}
\label{sec:large_vocabulary}

Tables~\ref{table:lvis} and~\ref{table:objects365} report object detection results on LVIS~\cite{gupta2019lvis} and Objects365~\cite{shao2019objects365}, respectively.
CenterNet2 improves on the CascadeRCNN baselines by 2.7 mAP on LVIS and 0.8 mAP on Objects365, showing the generality of our approach.
On both datasets, two-stage detectors (CascadeRCNN, CenterNet2) outperform one-stage designs (GFL, CenterNet) by significant margins: 5-8 mAP on LVIS and 3-4 mAP on Objects365.  
On LVIS, the runtime of one-stage detectors increases by $\sim\!30\%$ compared to COCO, as the number of categories grows from 80 to 1203. This is due to the dense classification heads.
On the other hand, the runtime of \name only increases by $5\%$.
This highlights the advantages of probabilistic two-stage detection in large-vocabulary settings.

Two stage-detectors allow using a more dedicated classification loss in the second stage. In the supplement, we propose a federated loss for handling the federated construction of LVIS. The results are highlighted in Table~\ref{table:lvis}.

\section{Conclusion}

We developed a probabilistic interpretation of two-stage detection.
This interpretation motivates the use of a strong first stage that learns to estimate object likelihoods rather than maximize recall. 
These likelihoods are then combined with the classification scores from the second stage to yield principled probabilistic scores for the final detections.
Probabilistic two-stage detectors are both faster and more accurate than their one- or two-stage counterparts.
Our work paves the way for an integration of advances in both one- and two-stage designs
that combines accuracy with speed.

\nocite{langley00}

\spacing{0.9}
\bibliography{example_paper}
\bibliographystyle{icml2021}
\normalsize
\spacing{1.0}
\clearpage
\appendix

\section{Tightness of lower bounds}

We briefly show that the max of the two lower bounds on the maximum likelihood objective is  indeed quite tight.
Recall the original training objective
\begin{equation*}
  \log P(bg) = \log \left(\underbrace{P(bg|O_k\!=\!1)}_\beta \underbrace{P(O_k\!=\!1)}_{1-\alpha} + \underbrace{P(O_k\!=\!0)}_{\alpha}\right).
\end{equation*}
We optimize two lower bounds
\begin{equation}
  \log P(bg) \ge \underbrace{\log \left(P(O_k\!=\!0)\right)}_{B_1}.
\end{equation}
and
\begin{equation}
  \log P(bg) \ge \underbrace{P(O_k\!=\!1) \log \left(P(bg|O_k\!=\!1)\right)}_{B_2}
\end{equation}
The combined bound is
\begin{equation}
  \log P(bg) \ge \max(B_1, B_2).
\end{equation}
This combined bound is within $\log(2)$ of the overall objective:
\begin{equation*}
  \log P(bg) \le \max(B_1, B_2) + \log(2).
\end{equation*}

We start by simplifying the max operation when computing the gap between bound and true objective:
\begin{align*}
  &\log P(bg) - \max(B_1, B_2)\\
  &=\begin{cases}\log P(bg) - B_1 & \text{if }B_1 \ge B_2\\\log P(bg) - B_2 & \text{otherwise}\end{cases}\\&\le \begin{cases}\log P(bg) - B_1 & \text{if }P(O_k=0) \ge P(bg|O_k=1)\\\log P(bg) - B_2 & \text{otherwise}\end{cases}.
\end{align*}
Here the last inequality holds by definition of the max ($-\max(a,b) \le -a$ and $-\max(a,b) \le -b$).
Let us analyze each case separately.

\paragraph{Case 1: $P(O_k=0) \ge P(bg|O_k=1)$ i.e. $\alpha \ge \beta$}

Here, we analyze the bound
\begin{align*}
  \log P(bg) - B_1 &= \log(\beta(1-\alpha)+\alpha) - \log \alpha\\
  \intertext{Due to the monotonicity of the $\log$ the maximal value of the above expression for $\beta \le \alpha$ and $1-\alpha \ge 0$ is the largest possibble values $\beta=\alpha$. Hence for any value $\beta \le \alpha$ :}
  \log P(bg) - B_1 &\le \log(\alpha(1-\alpha)+\alpha) - \log \alpha\\
  &= \log(2-\alpha) \le \log(2),\\
\end{align*}
since $\alpha \ge 0$.

\paragraph{Case 2: $P(O_k=0) \le P(bg|O_k=1)$ i.e. $\alpha \le \beta$}

Here, we analyze the bound
\begin{align*}
  \log P(bg) - B_2 &= \log(\beta(1-\alpha)+\alpha) - (1-\alpha)\log \beta\\
  \intertext{since $(1-\alpha) \le 1$:}
  \log P(bg) - B_2 &\le \log(\beta(1-\alpha)+\alpha) - \log \beta\\
  \intertext{since $\alpha \le \beta$ the above is maximal at the largest values $\alpha \le \beta$ since $\beta \le 1$ (hence at $\alpha = \beta$ again):}
  \log P(bg) - B_2 &\le \log(2-\beta) \le \log 2\\
\end{align*}

Both parts of the max-bound come within $\log 2$ of the actual objective. Most interestingly they are exactly $\log 2$ away only at $\alpha = \beta \to 0$, where the objective value $\log(\beta(1-\alpha)+\alpha) \to -\infty$ is at negative infinity, and are tighter than $\log 2$ for all other values.

\section{Backbones and training details}

\paragraph{Default backbone.}
We implement our method based on detectron2~\cite{wu2019detectron2}.
Our default model follows the standatd Res50-1x setting in detectron2~\cite{wu2019detectron2}.
Specifically, we use ResNet-50~\cite{he2016deep} as the backbone, and train the network with the SGD optimizer for 90K iterations (1x schedule). 
The base learning rate is $0.02$ for two-stage detectors and $0.01$ for one-stage detectors, 
and is dropped by 10x at iterations 60K and 80K.
We use multi-scale training with the short edge in the range [640,800] and the long edge up to 1333.
During training, we set the first-stage loss weight to $0.5$ as one-stage detectors are typically trained with learning rate $0.01$.
During testing, we use a fixed short edge at 800 and long edge up to 1333.

\begin{table*}[!t]
\centering
\begin{tabular}{@{}l@{\ \ \ }l@{\ \ }c@{\ \ \ \ }c@{\ \ \ \ }c@{\ \ \ \ }c@{}}
\toprule
Detector & Loss & $AP^{box}$ & $AP^{box}_{r}$ & $AP^{box}_{c}$ & $AP^{box}_{f}$ \\
\midrule
\multirow{4}{7em}{CenterNet2}
& Softmax-CE & $26.9$ \scriptsize $ \pm 0.05$ & $12.4$ \scriptsize $ \pm 0.15$ & $25.4$ \scriptsize $ \pm 0.15$ & $35.0$ \scriptsize $ \pm 0.05$  \\
& Sigmoid-CE & $26.6$ \scriptsize $ \pm 0.00$ & $12.4$ \scriptsize $ \pm 0.10$ & $25.1$ \scriptsize $ \pm 0.05$ & $34.5$ \scriptsize $ \pm 0.10$ \\
& EQL~\cite{tan2020eql} & $27.3$ \scriptsize $ \pm 0.00$ & $15.1$ \scriptsize $ \pm 0.45$ & $25.9$ \scriptsize $ \pm 0.20$ & $34.2$ \scriptsize $ \pm 0.35$ \\
& FedLoss (Ours) & $\bf 28.2$ \scriptsize $ \pm 0.05$ & $\bf 18.8$ \scriptsize $ \pm 0.05$ & $\bf 26.4$ \scriptsize $ \pm 0.00$ & $\bf 34.4$ \scriptsize $ \pm 0.00$\\
\midrule
\multirow{4}{7em}{CascadeRCNN}
& Softmax-CE  & $24.0$ \scriptsize $ \pm 0.10$ & $7.6$ \scriptsize $ \pm 0.10$ & $22.9$ \scriptsize $ \pm 0.15$ & $32.7$ \scriptsize $ \pm 0.05$  \\
& Sigmoid-CE & $23.3$ \scriptsize $ \pm 0.10 $ & $8.2$ \scriptsize $ \pm 0.30$ & $21.9$ \scriptsize $ \pm 0.40$ & $31.5$ \scriptsize $ \pm 0.25$ \\
& EQL~\cite{tan2020eql} & $25.7$ \scriptsize $ \pm 0.01$ & $15.5$ \scriptsize $ \pm 0.25$ & $24.6$ \scriptsize $ \pm 0.70$ & $31.5$ \scriptsize $ \pm 0.45$ \\
& FedLoss (Ours) & $\bf 27.1$ \scriptsize $ \pm 0.05$ & $\bf 16.1$ \scriptsize $ \pm 0.10$ & $\bf 26.0$ \scriptsize $ \pm 0.35$ & $\bf 33.0$ \scriptsize $ \pm 0.25$ \\
\bottomrule
\end{tabular}
\normalsize
\caption{Ablation experiments on different classification losses on LVIS v1 validation. We show results with both our proposed detector (top) and the baseline detector (bottom). All models are ResNet50-1x with FPN P3-P7 and multi-scale training. We report mean and standard deviation over 2 runs.}
\label{table:fedloss}
\vspace{-3mm}
\end{table*}

\paragraph{Large backbone.}
Following recent works~\cite{tian2019fcos,zhu2020autoassign,zhang2020bridging},
we use ResNeXt-32x8d-101-DCN~\cite{xie2017aggregated} as a large backbone.
Deformable convolutions~\cite{zhu2019deformable} are added to Res4-Res5 layers.
It is trained with a 2x schedule (180K iterations with the learning rate dropped at 120K and 160K). 
We extend the scale augmentation to set the shorter edge in the range [480,960] for the large model~\cite{zhang2019freeanchor,zhu2020autoassign,chen2020reppointsv2}.
The test scale is fixed at 800 for the short edge and up to 1333 for the long edge.

\noindent \textbf{Real-time backbone.}
We follow real-time FCOS~\cite{tian2020fcos} and use DLA~\cite{yu2018deep} with BiFPN~\cite{tan2020efficientdet}.
We use 4 BiFPN layers with feature channels 160~\cite{tian2020fcos}.
The output FPN levels are reduced to 3 levels with stride 8-32.
We train our model with scale augmentation and set the short edge in the range [256,608], with the long edge up to 900.
We first train with a 4x schedule (360K iterations with the learning rate dropped at the last 60K and 20K iterations) to compare with real-time FCOS~\cite{tian2020fcos}.
We then train with a long schedule that repeatedly fine-tunes the model with the 4x schedule for 6 cycles (i.e., a total of 288 epochs).
During testing, we set the short edge at 512 and the long edge up to 736~\cite{tian2020fcos}.
We reduce the number of proposals to 128 for the second stage.
Other hyperparameters are unchanged.

\section{Extra-large model details}
To push the state-of-the-art results for object detection, we integrate recent advances into our framework to design an extra-large model.
Table.\ref{table:extralarge} outlines our changes.
Unless specified, we keep the test resolution as $800\times1333$ even when the training size changes.
We first switch the network backbone from ResNeXt-101-DCN to Res2Net-101-DCN~\cite{gao2019res2net}. This speeds up training, and gives a decent $0.6$ mAP improvement.
Next, we change the data augmentation style from the Faster RCNN style~\cite{wu2019detectron2,mmdetection} (random resize short edge) to EfficientDet style~\cite{tan2020efficientdet}, which involves resizing the original image and crop a square region from it. 
We first use a crop size of $896\times896$, which is close to the original $800\times1333$. 
We use a large resizing range of [0.1, 2] following the implementation in \citet{wightman2020}, and train with a $4\times$ schedule (360k iterations).
The stronger augmentation and longer schedule together improve the result to 51.4 mAP.
Next, we change the crop size to $1280\times1280$, and add BiFPN~\cite{tan2020efficientdet} to the backbone. We follow EfficientDet~\cite{tan2020efficientdet} to use 288 channels and 7 layers in the BiFPN that fits the $1280\times1280$ input size.
This brings the performance to 53.0 mAP.
Finally, we use the technics in ~\citet{zoph2020rethinking} to use COCO unlabeled images~\cite{lin2014coco} for self-training.
Specifically, we run a YOLOv4~\cite{wang2020scaled} model on the COCO unlabeled images. 
We set all predictions with scores $>0.5$ as pseudo-labels.
We then concatenate this new data with the original COCO training set, and finetune our previous best model on the concatenated dataset for another $4\times$ schedule.
This model gives 54.4 mAP with test size $800\times1333$.
When we increase the test size to $1560\times1560$, the performance further raises to $56.1$ mAP on COCO validation and $56.4$ mAP on COCO test-dev.

\begin{table}[!t]
\centering
\begin{tabular}{l@{\ \ \ \ \ \ }c}
\toprule
 & mAP \\
\midrule
CenterNet2 & 49.9 \\
+ Res2Net-101-DCN & 50.6 \\
+ Square crop Aug \& 4x schedule & 51.2 \\
+ train size $1280\!\times\!1280$ \& BiFPN & 52.9 \\
+ ft. w/ self training 4x & 54.4 \\
+ test size $1560\!\times\!1560$ & 56.1 \\
\bottomrule
\end{tabular}
\vspace{-2mm}
\caption{Road map from the large backbone to the extra-large backbone. We show COCO validation mAP. }
\label{table:extralarge}
\vspace{-5mm}
\end{table}

We also combined part of these advanced training technics in our real-time model.
Specifically, we use the EfficientDet-style square-crop augmentation, use the original FPN level P3-P7 (instead of P3-P5), and use self-training.
These modifications improves our real-time model from $45.6$mAP@ $25$ms to $49.2$ mAP@ $30$ms.

\section{Federated Loss for LVIS}

LVIS annotates images in a federated way~\cite{gupta2019lvis}.
I.e., the images are only sparsely annotated.
This leads to much sparser gradients, especially for rare classes~\cite{tan2020eql}.
On one hand, if we treat all unannotated objects as negatives, the resulting detector will be too pessimistic and ignore rare classes.
On the other hand, if we only apply losses to annotated images the resulting classifier will not learn a sufficiently strong background model.
Furthermore, neither strategy reflects the natural distribution of positive and negative labels on a potential test set.
To remedy this, we choose a middle ground and apply a \emph{federated loss} to a subset $S$ of classes for each training image.
$S$ contains all positive annotations, but only a random subset of negatives.

We sample the negative categories in proportion to their square-root frequency in the training set, and empirically set $|S| = 50$ in our experiments.
During training, we use a binary cross-entropy loss on all classes in $S$ and ignore classes outside of $S$.
The set $S$ is sampled per iteration.
The same training image may be in different subsets of classes in consecutive iterations.

Table~\ref{table:fedloss} compares the proposed federated loss to baselines including the LVIS v0.5 challenge winner, the equalization loss (EQL)~\cite{tan2020eql}.
For EQL, we follow the authors' settings in LVIS v0.5 to ignore the 900 tail categories.
Switching from the default softmax to sigmoid incurs a slight performance drop.
However, our federated loss more than makes up for this drop, and outperforms EQL and other baselines significantly.

\section{Comparison with other proposal networks}

\begin{table}[t]
\centering
\begin{tabular}{l@{\ \ \ \ }c@{\ \ \ \ }c}
\toprule
 & mAP & Runtime\\
\midrule
FasterRCNN-CenterNet & 40.8 & 50ms\\
GA RPN~\cite{wang2019region} & 39.6 & 75ms \\
Cascade RPN~\cite{vu2019cascade} & 40.4 & 97ms \\
\bottomrule
\end{tabular}
\vspace{-2mm}
\caption{Comparison to other proposal networks. All models are trained with Res50-1x without data augmentation. The models of GA RPN and Cascade RPN are from mmdetection~\cite{mmdetection}}
\label{table:otherrpn}
\vspace{-5mm}
\end{table}

GA RPN~\cite{wang2019region} and CascadeRPN~\cite{vu2019cascade} also improves the original RPN, by using deformable convolutions~\cite{zhu2018deformable} in the RPN layers~\cite{wang2019region} or using a cascade of proposal networks~\cite{vu2019cascade}.
We train a probablistic two-stage detector FasterRCNN-CenterNet under the same setting (FasterRCNN, Res50-1x, without data augmentation), 
and compare them in Table ~\ref{table:otherrpn}.
Our model performs better than both proposal network, and runs faster.

{
\section{Dataset details}
We use the official release and the standard train/ validation splot.
COCO~\cite{lin2014coco} contains 118k training images, 5k validation images, and 20k test images for 80 categories.
LVIS (V1)~\cite{gupta2019lvis} contains 100k training images and 20k validation images for 1203 categories. 
Objects365~\cite{shao2019objects365} contains 600k training images and 30k validation images for 365 categories.
All datasets collect images from the internet and provide accurate annotations.
}

\end{document}